\documentclass[journal]{IEEEtran}
\usepackage{amsmath,amsfonts}
\usepackage{bm}
\usepackage{breqn}
\usepackage{booktabs}    
\usepackage{makecell}    
\usepackage{adjustbox}   
\usepackage[normalem]{ulem}        
\usepackage{multirow}    
\usepackage{tabularx}    
\usepackage{graphicx}  
\usepackage{algorithmic}
\usepackage{algorithm}
\usepackage{array}
\usepackage[caption=false,font=normalsize,labelfont=sf,textfont=sf]{subfig}
\usepackage{textcomp}
\usepackage{stfloats}
\usepackage{url}
\usepackage{verbatim}
\usepackage{cite}
\usepackage{orcidlink}
\hypersetup{
    colorlinks=true,
    linkcolor=blue,
    citecolor=blue,
    urlcolor=blue,
    breaklinks=true
}
\begin{document}

\title{RAFNet: Region-Aware Fusion Network for Pansharpening}

\author{Jianing Zhang$^{\orcidlink{0009-0007-5358-8935}}$, Zijian Zhou$^{\orcidlink{0009-0003-9505-4086}}$, Kai Sun$^{\orcidlink{0000-0003-2635-5478}}$
\thanks{Manuscript received [Month Day, Year]; revised [Month Day, Year].}
\thanks{This work was supported in part by the National
Natural Science Foundation of China under Grant 12201490, in part by the China Postdoctoral Science Foundation under Grant 2025M773093, and in part by the Postdoctoral Fellowship Program of CPSF under Grant GZC20252028. \textit{(Corresponding author: Kai Sun.)}}
\thanks{The first two authors contributed equally.}
\thanks{Jianing Zhang and Zijian Zhou are with the College of Artificial Intelligence, Xi’an Jiaotong University, Xi’an 710049, China (e-mail: 2518720025@stu.xjtu.edu.cn; zzj210119@stu.xjtu.edu.cn).}
\thanks{Kai Sun is with the School of Mathematics and Statistics, Xi’an Jiaotong University, Xi’an 710049, China (e-mail: kaisun@mail.xjtu.edu.cn).}
}

\markboth{IEEE Transactions on Geoscience and Remote Sensing,~Vol.~XX, No.~XX, [Month]~2026}%
{Authors \MakeLowercase{\textit{et al.}}: RAFNet for Pansharpening}

\IEEEpubid{
\begin{minipage}{\textwidth}
    \centering
    \vspace{40pt} 
    1558-0644 \copyright~2026 IEEE. All rights reserved, including rights for text and data mining, and training of artificial intelligence \\
    and similar technologies. Personal use is permitted, but republication/redistribution requires IEEE permission.\\
    See https://www.ieee.org/publicatios/rights/index.html for more information.
\end{minipage}
}

\maketitle

\begin{abstract}
Pansharpening aims to generate high-resolution multispectral (HRMS) images by fusing low-resolution multispectral (LRMS) and high-resolution panchromatic (PAN) images. Although deep learning has advanced this field, mainstream frequency-based methods relying on standard scaled dot-product attention suffer from quadratic computational complexity and fail to exploit the inherent regional sparsity of remote sensing imagery. Furthermore, existing spatial enhancement strategies typically employ static convolution kernels, which struggle to adapt to the complex frequency and regional variations of PAN and MS images. To address these bottlenecks, we propose a Region-Aware Fusion (RAFNet) Network that synergistically models spatial and frequency information. Specifically, we design a Spatial Adaptive Refinement (SAR) module that leverages the discrete wavelet transform (DWT) for directional frequency separation and K-means clustering for regional partitioning, which enables the dynamic construction of region-specific adaptive convolution kernels, achieving spatially and frequency-adaptive feature enhancement. Moreover, we introduce a Clustered Frequency Aggregation  (CFA) module based on a sparse attention mechanism guided by the semantic clusters, which executes a region-aware sparse attention strategy that drastically reduces computational redundancy while ensuring high-quality frequency feature extraction. In addition we integrated these modules into a progressive, multi-level spatial-frequency network architecture to facilitate robust interaction and accurate image reconstruction. Extensive experiments on multiple benchmark datasets demonstrate that the proposed RAFNet significantly outperforms state-of-the-art pansharpening methods in both reduced- and full-resolution assessments. The code is available at https://github.com/PatrickNod/RAFNet.
\end{abstract}


\section{Introduction}

\IEEEPARstart{T}{he} unprecedented demand for high-resolution multispectral (HRMS) imagery in Earth observation \cite{1, 2} is hindered by the inherent physical trade-offs of optical sensors. Consequently, satellites typically acquire low-resolution multispectral (LRMS) and high-resolution panchromatic (PAN) images separately. While LRMS preserves rich spectral fidelity, PAN captures fine spatial details. Pansharpening has emerged as a crucial paradigm to circumvent this limitation, aiming to synergistically fuse the spectral information of LRMS with the spatial structures of PAN to reconstruct high-quality HRMS imagery.

Traditional pansharpening methods are broadly categorized into three classes \cite{3}: Component Substitution (CS) \cite{4, 5}, Multi-Resolution Analysis (MRA) \cite{6, 7}, and Variational Optimization (VO) \cite{8, 9}. They frequently provoke severe spectral distortion \cite{10}, introduce spatial artifacts like ringing \cite{11}, or rely on hand-crafted priors that fail to model complex real-world remote sensing data \cite{12}.

\IEEEpubidadjcol

In recent years, deep learning architectures have achieved remarkable success in remote sensing image fusion \cite{13, 14}. Early paradigms directly fed stacked PAN and MS images into convolutional networks for end-to-end mapping \cite{15}. Subsequently, methods that inject neural-extracted spatial details into upsampled MS images substantially improved fusion fidelity \cite{16, 17}. Representative architectures like U2Net \cite{18} employ multi-scale networks to capture and enhance these details. However, relying exclusively on static convolutional kernels, these methods inherently lack spatial adaptability and struggle to model diverse regional distributions across input images \cite{19}.

\begin{figure*}[!t]
\centering
\includegraphics[width=1\textwidth]{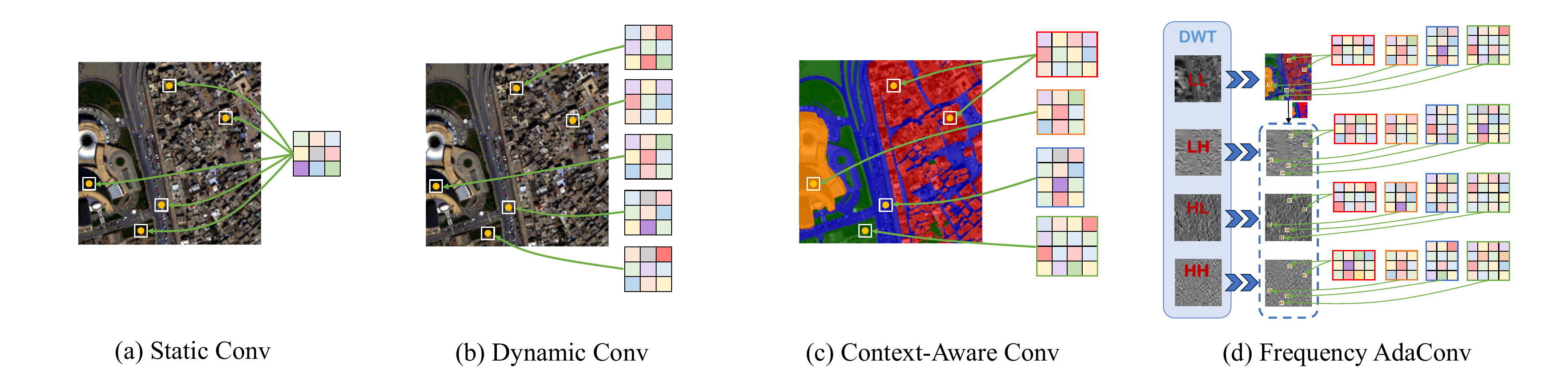} 
\caption{Evolution of convolution paradigms for spatial feature enhancement. (a) Static Conv applies a spatially invariant filter across the entire image, often attenuating high-frequency details. (b) Dynamic Conv generates localized, spatially adaptive weights but fails to exploit non-local similarities. (c) Context-Aware Conv leverages regional clustering to assign shared adaptive kernels to semantically similar non-local regions. (d) Frequency AdaConv integrates the discrete wavelet transform (DWT) to synthesize region-adaptive dynamic convolution kernels specifically tailored to distinct high-frequency components (LH, HL, HH) besides the low-frequency (LL) guided by the low-frequency (LL) semantic mask.}
\label{fig:fig1}
\end{figure*}

Furthermore, Yedla et al. \cite{20} revealed a spectral bias where conventional networks over-focus on low-frequency content at the expense of high-frequency details. Since intricate spatial details are embedded within high-frequency components \cite{21}, static convolutions inadvertently attenuate them, as illustrated in Fig.~\ref{fig:fig1}(a). While some studies employed linear transformations to mitigate frequency entanglement \cite{22}, Multilayer Perceptron (MLP) modules disrupt the intrinsic spatial topology of pixels \cite{23}. Consequently, the paradigm shifted toward spatially adaptive convolutions. Early methods like DFN \cite{24} and DDF \cite{25} synthesized localized dynamic kernels conditioned on input content (Fig.~\ref{fig:fig1}(b)), but failed to exploit non-local self-similarities. Motivated by regional correlations, methods like IGNN \cite{27} and Canconv \cite{28} integrated clustering mechanisms to extract non-local adaptive priors (Fig.~\ref{fig:fig1}(c)), significantly bolstering spatial adaptability.

Although cluster-guided convolutions present a promising paradigm, Chen et al. \cite{29} demonstrated that the frequency-domain responses of dynamically generated weights often exhibit high similarity. To alleviate this, they proposed FDconv to construct adaptive kernels within the Fourier domain, underscoring that optimal kernel generation must simultaneously account for spatial adaptability and frequency variations. This indicates existing strategies remain inadequately aligned with the complex frequency distributions of remote sensing data. Building upon these critical insights, our proposed method focuses on synthesizing region-adaptive dynamic convolution kernels specifically tailored to distinct frequency components via DWT (Fig.~\ref{fig:fig1}(d)), achieving a superior and highly generalized spatial enhancement paradigm.

\begin{figure*}[!t]
\centering
\includegraphics[width=0.85\textwidth]{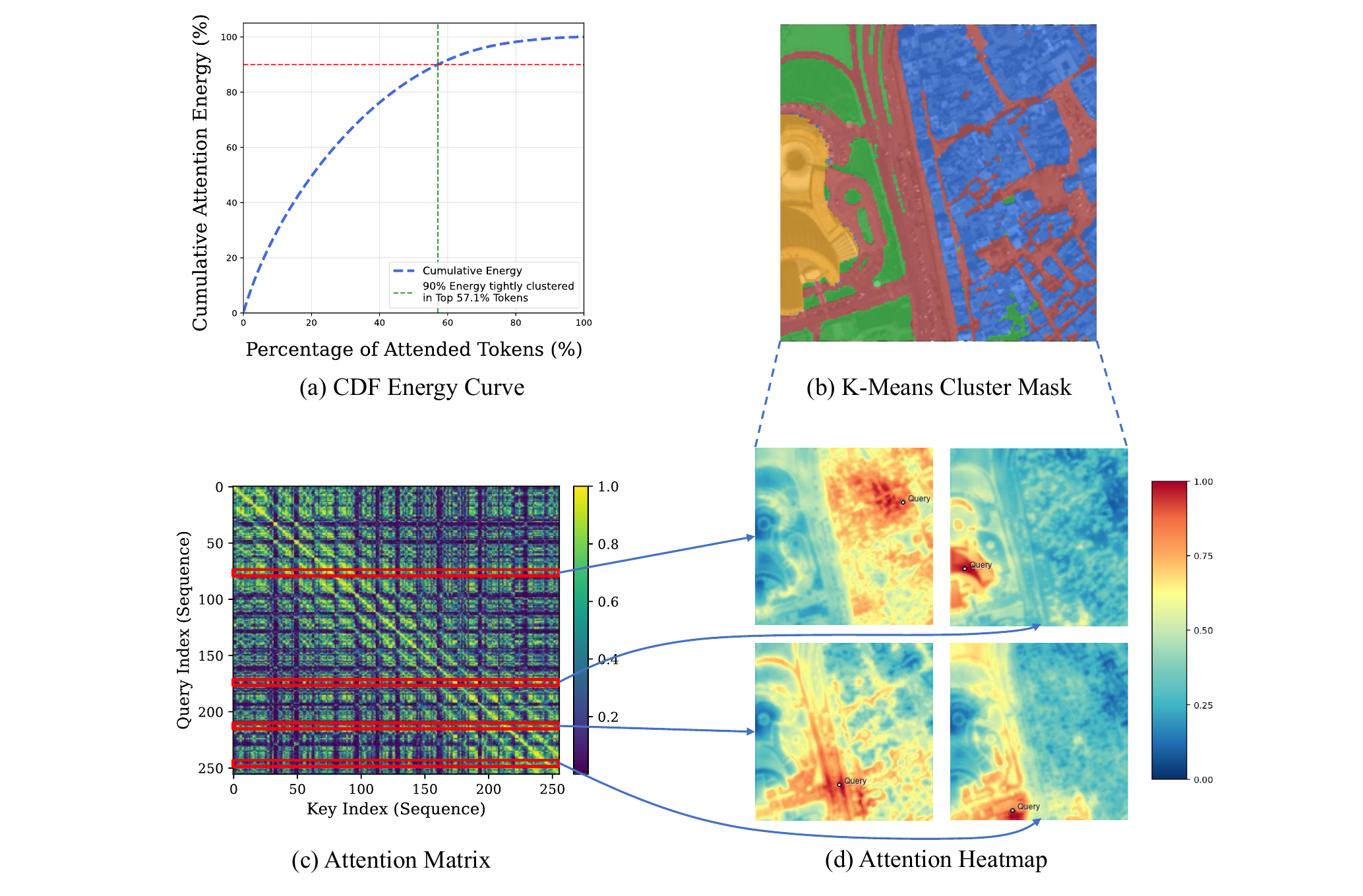} 
\caption{Motivation for the proposed Cluster-Routed Sparse Attention (CRSA). (a) CDF Energy Curve: Demonstrates the long-tail distribution of attention weights, where 90\% of the energy is tightly clustered within only the top 57.1\% of tokens. (b) K-Means Cluster Mask: Illustrates the spatial semantic regions of the image. (c) Attention Matrix: Visualization of the global attention matrix, revealing severe block-diagonal and sparse structures with massive inactive interactions. (d) Attention Heatmap: Spatial attention maps corresponding to the specific query indices highlighted by red boxes in (c). The heatmaps project the attention weights back into the 2D spatial domain, with warmer colors indicating higher attention scores. Crucially, the boundaries of the high-attention regions (red/yellow) for each query pixel align with the shape of its corresponding semantic cluster shown in (b). This visually confirms the spatial semantic alignment phenomenon, providing a strong empirical rationale for using regional clustering to guide sparse attention interactions.}
\label{fig:fig2}
\end{figure*}

Furthermore, relying exclusively on the injection of spatial details is often suboptimal for comprehensive pansharpening inference \cite{30}. Consequently, frequency-domain image fusion techniques have emerged as a superior alternative, with Transformer-based architectures rapidly establishing themselves as the dominant paradigm \cite{31, 32}. Nevertheless, the deployment of these attention-based models in practical remote sensing scenarios is severely hindered by inherent technical bottlenecks. The computational time and memory footprint of standard attention mechanisms scale quadratically with the spatial sequence length, which consumes prohibitive computational resources and introduces substantial information redundancy \cite{33}. To empirically validate this redundancy, we analyze the cumulative distribution function (CDF) energy curve in Fig.~\ref{fig:fig2}(a), which reveals an extreme long-tail distribution where 90\% of the useful attention energy is tightly clustered in a mere 57.1\% of the tokens. Furthermore, the visualization of the global attention matrix (Fig.~\ref{fig:fig2}(c)) reveals a pronounced block-diagonal and highly sparse structure, visually proving that the vast majority of query-key interactions compute near-zero weights and contribute minimally to the final representation. 

Therefore, mitigating the computational overhead of attention mechanisms while preserving representation capabilities remains a critical challenge. To address this, pioneering studies have proposed sparse approximations to constrain the number of active Query-Key interactions, or low-rank approximations to compress the correlation matrices \cite{34, 35}. Early sparse attention frameworks like CPC \cite{36} employed banded attention coupled with global node constraints to restrict Query interactions to local neighborhoods to achieve linear complexity. Similarly, the REFORMER \cite{37} utilized Locality-Sensitive Hashing (LSH) to probabilistically select relevant Key-Value pairs. However, these methods typically impose spatially invariant attention patterns across the entire image, failing to accommodate the diverse semantic regions inherent in complex scenes. The Routing Transformer \cite{38} alleviated this rigidity by leveraging K-means clustering, restricting each Query to exclusively attend to Keys within its assigned semantic cluster.

In the specialized domain of remote sensing pansharpening, sparse attention paradigms are gradually gaining traction. For instance, PMRF \cite{39} adopted Swin-Transformer architecture to explicitly curtail computational complexity while maintaining fusion fidelity, but the spatial flexibility of its static window partitioning remains fundamentally limited. This underscores a pressing imperative to design customized sparse attention mechanisms tailored to the regional semantic features of remote sensing data. To guide this design, we further investigate the spatial distribution of the active attention energy. As graphically depicted by comparing the K-Means cluster mask (Fig.~\ref{fig:fig2}(b)) with the spatial attention heatmaps of selected query pixels (Fig.~\ref{fig:fig2}(d)), a striking phenomenon of spatial semantic alignment emerges. The attention energy naturally concentrates strictly within the same semantic boundaries. Motivated by these critical insights into global matrix sparsity and spatial semantic alignment, our proposed method integrates the advantages of spatial clustering to formulate a highly efficient, region-adaptive sparse attention mechanism, effectively resolving the dilemma between computational complexity and fusion performance.

To surmount the aforementioned limitations in spatial enhancement and frequency-domain fusion, we propose a novel Region-Aware Fusion (RAFNet) Network. Specifically, we design a Spatial Adaptive Refinement (SAR) module grounded in the discrete wavelet transform (DWT) for precise directional frequency decoupling. By employing K-means clustering to dynamically synthesize region-specific adaptive convolution kernels, the SAR explicitly captures the correlation between frequency distributions and spatial features to robustly enhance high-frequency details.

Furthermore, to mitigate the prohibitive computational overhead of standard global attention, we introduce the Clustered Frequency Aggregation (CFA) module. By exploiting the inherent regional sparsity of attention matrices and integrating the semantic guidance from the SAR, the CFA executes a region-aware sparse attention strategy. This substantially reduces complexity while guaranteeing the unimpeded propagation of fine-grained details and global image structures.

Finally, we embed these modules within a multi-scale architecture tailored to the unique characteristics of remote sensing imagery. By strategically harmonizing DWT, K-means clustering, adaptive convolutions, and sparse attention, our framework rigorously prunes computational redundancy. This achieves an optimal trade-off between computational efficiency and fusion performance, ensuring superior HRMS image reconstruction.

In summary, our main contributions are as follows:
\begin{enumerate}
    \item We propose a novel Region-Aware Fusion (RAF) Network for pansharpening that synergizes spatial enhancement with frequency-domain fusion. By facilitating collaborative spatial-spectral processing, RAFNet circumvents the limitations of existing architectures to achieve state-of-the-art performance, while its modular design offers a highly generalized paradigm for remote sensing image fusion.
    \item We design a Spatial Adaptive Refinement (SAR) module that integrates the discrete wavelet transform with K-means semantic clustering. By coupling directional frequency decoupling with region-specific adaptive convolution kernels, the SAR module ensures the precise extraction and high-fidelity propagation of complex spatial details.
    \item We introduce a Cluster-Routed Sparse Attention (CRSA) module that leverages semantic clustering priors from SAR to adaptively route the flow of frequency information. This region-aware mechanism radically mitigates the quadratic computational complexity of standard attention, providing a highly efficient global fusion strategy optimized for high-resolution imagery.
\end{enumerate}

\begin{figure*}[!t]
\centering
\includegraphics[width=0.95\textwidth]{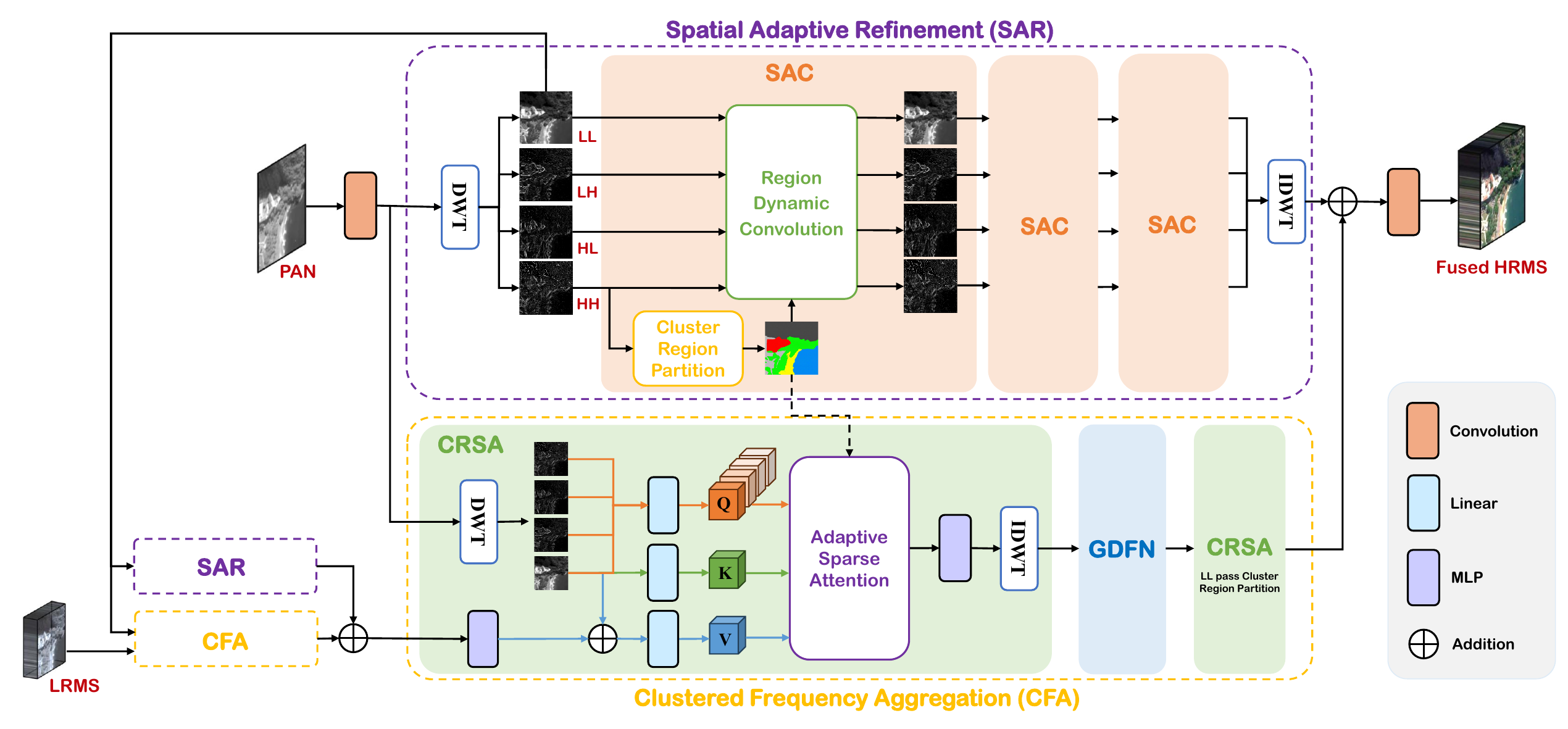} 
\caption{Overall architecture of the proposed Region-Aware Fusion (RAF) network. The network takes PAN and LRMS images as inputs and primarily consists of two core modules. The Spatial Adaptive Refinement (SAR) extracts and enhances spatial details via Discrete Wavelet Transform (DWT) and cascaded Spatial Adaptive Convolutions (SACs). The Clustered Frequency Aggregation (CFA) fuses information across frequencies using adaptive sparse attention and Gated Depthwise Forward Networks (GDFN). The final high-resolution multispectral (HRMS) image is reconstructed progressively.}
\label{fig:overall_arch}
\end{figure*}

\section{Related Work}

\subsection{Dynamic Convolution Based Pansharpening}
Adaptive convolutions dynamically synthesize content-conditioned kernels to improve spatial representation over standard static weights. While early methods like DFN \cite{24}, DRConv \cite{41}, and DDF \cite{25} introduced increasingly granular, pixel-specific kernels to address spatial variance, they often incur severe computational redundancy and apply spatially invariant parameters that fail to capture diverse regional heterogeneity \cite{40}. Furthermore, these spatial-centric techniques inherently neglect image frequency priors, leading to a spectral bias that suppresses crucial high-frequency details in favor of low-frequency structures \cite{29}. Although recent remote sensing applications like LAGConv \cite{26} and Canconv \cite{28} have successfully leveraged local and non-local similarities to augment spatial fidelity, they still struggle with this spectral imbalance. Motivated by these limitations, we design a specialized adaptive convolution module that synergistically integrates rich frequency priors with spatial adaptability to prevent the attenuation of high-frequency spatial details.

\subsection{Sparse Attention Based Pansharpening}
Transformer architectures excel at capturing long-range global dependencies, making them highly efficacious for pansharpening applications \cite{42, 43, 44, 45}. However, their standard scaled dot-product attention incurs a prohibitive quadratic computational and memory complexity. To alleviate this, sparse attention mechanisms have been extensively investigated to balance representation performance and efficiency \cite{33}. Methods such as banded attention \cite{36}, Locality-Sensitive Hashing (LSH) \cite{37}, and transposed attention mechanisms \cite{46} successfully reduce computational overhead. Nevertheless, these conventional sparse approaches typically impose rigid, spatially invariant attention patterns, lacking the adaptive flexibility required for complex visual scenes. To address this rigidity, the Routing Transformer \cite{38} introduced spatial adaptability by constraining interactions strictly within K-means semantic clusters. Concurrently, within remote sensing, architectures like PMRF \cite{39} have deployed window-based sparse paradigms to maintain fusion fidelity efficiently. Building upon these critical advancements, we propose a customized sparse attention module explicitly tailored to the inherent regional heterogeneity and high-resolution processing demands of remote sensing imagery.

\section{Proposed Method}

This section elaborates on the overall architectural design of the proposed Region-Aware Fusion (RAF) network. As illustrated in Fig. \ref{fig:overall_arch}, the RAF network is mainly composed of two core modules: the Spatial Adaptive Refinement (SAR) and the Clustered Frequency Aggregation (CFA). 

Let the input PAN image be $\bm{I}_P\in \mathbb{R}^{H\times W\times1}$, the low-resolution MS image be $\bm{I}_M\in \mathbb{R}^{\left(H/r\right)\times\left(W/r\right)\times C}$, and the ground-truth HRMS image be $\bm{I}_H\in \mathbb{R}^{H\times W\times C}$, where $r$ is the spatial resolution ratio and $C$ is the number of spectral channels. In typical pansharpening scenarios, $r=4$. Since a single DWT operation halves the spatial resolution, our RAF network naturally adopts a two-stage progressive architecture. To provide a clear explanation, the following subsections (\ref{subsec:sar} and \ref{subsec:cfa}) detail the internal mechanisms of the SAR and CFA at a generic $k$-th stage, where they process the intermediate panchromatic frequency features $\bm{P}_k$ and the intermediate multispectral features $\bm{M}_k$.

\begin{figure*}[!t]
\centering
\includegraphics[width=1\textwidth]{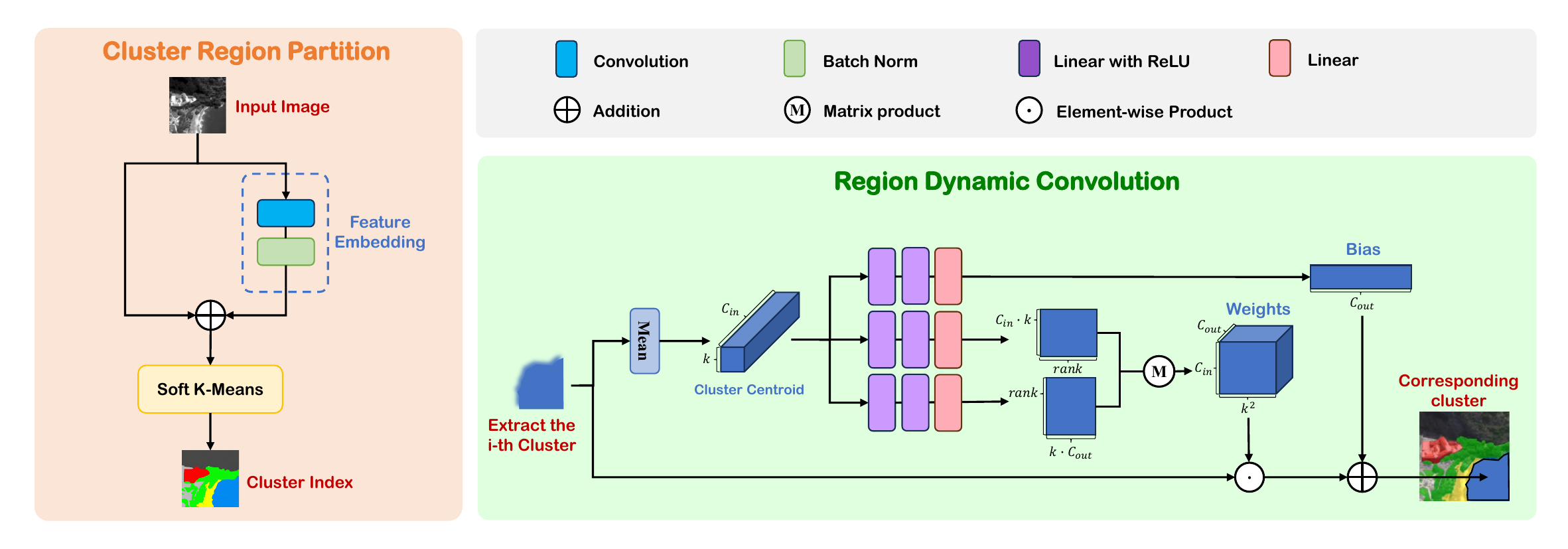} 
\caption{Detailed architectures the Spatial Adaptive Convolution (SAC) unit within the SAR module. It comprises the Cluster Region Partition for generating semantic masks via Soft K-Means, and the Region Dynamic Convolution for producing region-specific kernels using a low-rank MLP architecture.}
\label{fig:detailed_modules}
\end{figure*}

\subsection{SAR: Spatial Adaptive Refinement}\label{subsec:sar}
The SAR is designed to adaptively extract and enhance spatial detail information at each scale. It utilizes a 2D Discrete Wavelet Transform (DWT) followed by cascaded Spatial Adaptive Convolutions (SACs) and an Inverse DWT (IDWT) for lossless spatial reconstruction.

To effectively disentangle the spatial features into distinct frequency representations without computational overhead, we employ the standard 2D Haar DWT \cite{22}. It is elegantly formulated as depthwise convolutions utilizing four fixed orthogonal kernels (capturing low-frequency baseline and high-frequency residuals along horizontal, vertical, and diagonal directions), followed by a spatial downsampling operation.

At the $k$-th stage, let $\ast$ denote the 2D convolution operation and $\downarrow_2$ denote spatial downsampling with a stride of 2. The frequency decomposition is executed as:
\begin{equation}\label{eq:dwt_conv}
\bm{P}_{b} = (\bm{P}_k \ast \bm{f}_{b}) \downarrow_2, \quad b \in \{LL, LH, HL, HH\}
\end{equation}
where $\bm{P}_{LL}$ captures the down-sampled baseline spatial structure (low frequency), while $\bm{P}_{LH}, \bm{P}_{HL}$, and $\bm{P}_{HH}$ explicitly extract the high-frequency residual details along the horizontal, vertical, and diagonal directions, respectively. 

These decoupled sub-bands are subsequently fed into the SAC, which consists of a Cluster Region Partition and a Region Dynamic Convolution, as illustrated in Fig. \ref{fig:detailed_modules}.

\textit{1) Cluster Region Partition: }To capture non-local self-similarity, we perform a differentiable soft K-Means clustering \cite{56} on the low-frequency component $\bm{P}_{LL}$. For a pixel with feature embedding $\bm{f}_n$ and spatial coordinate $\bm{p}_n$, its soft assignment probability to the $k$-th cluster (with feature center $\bm{c}_k$ and spatial center $\bm{s}_k$) is computed via a temperature-scaled Softmax function:
\begin{equation}\label{eq:softmax_assignment}
\bm{S}_{n,k} = \frac{\exp(-(\left\| \bm{f}_n - \bm{c}_k \right\|_2 + \lambda \left\| \bm{p}_n - \bm{s}_k \right\|_2) / \tau)}{\sum_{j=1}^{K} \exp(-(\left\| \bm{f}_n - \bm{c}_j \right\|_2 + \lambda \left\| \bm{p}_n - \bm{s}_j \right\|_2) / \tau)}
\end{equation}
where $\lambda$ acts as a spatial penalty weight and $\tau$ is the temperature scaling factor.

The output is the discrete cluster index matrix $\bm{M} \in \mathbb{N}^{H \times W}$, acting as a regional mask. To ensure this argmax-based discrete assignment remains end-to-end differentiable during backpropagation, we employ the standard Straight-Through Estimator (STE) by forwarding the hard cluster assignments while bypassing gradients through the soft probabilities $\bm{S}_{n,k}$.

\begin{figure*}[!t]
\centering
\includegraphics[width=1\textwidth]{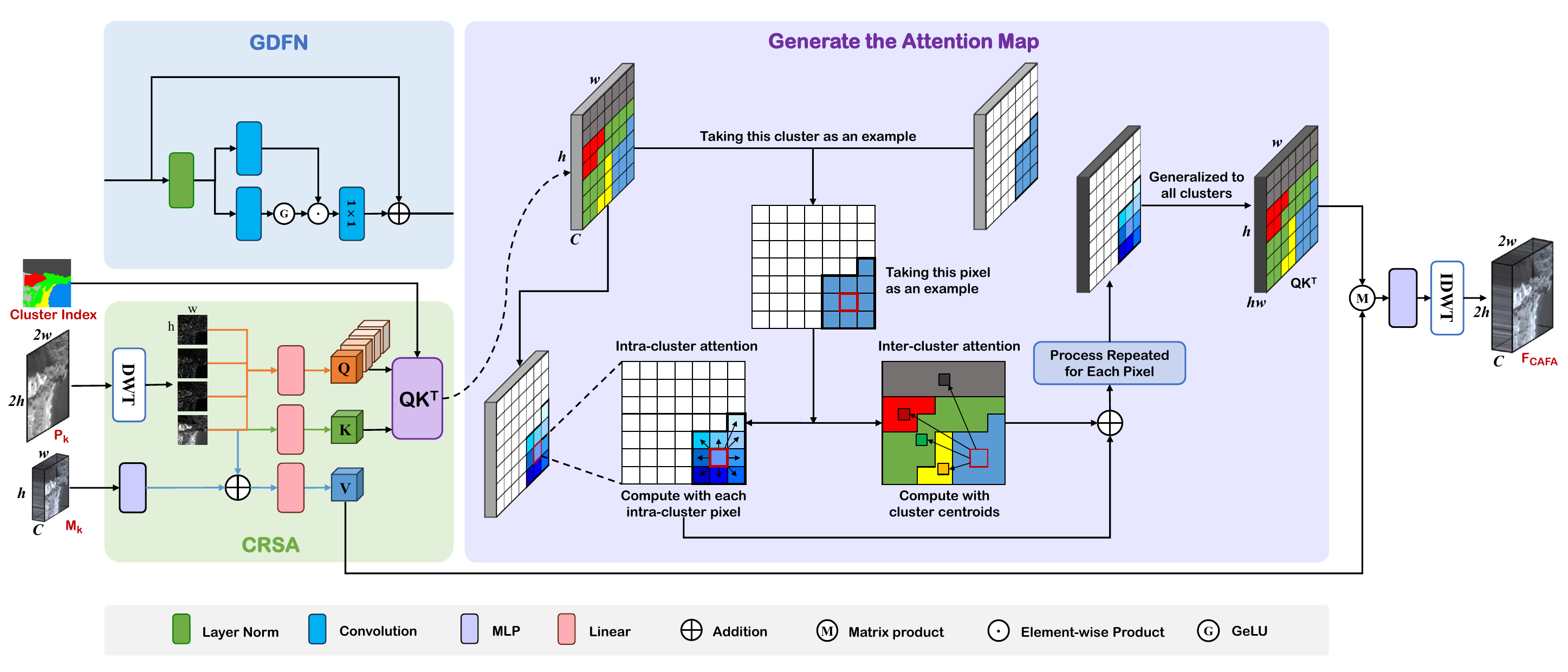} 
\caption{Architectures of the Clustered Frequency Aggregation (CFA) module, detailing the Cluster-Routed Sparse Attention (CRSA) and the GDFN. The right side of the bottom panel illustrates the sparse attention strategy, which performs dense dot-product for intra-cluster pixels and uses cluster-mean keys for inter-cluster interactions to significantly reduce computational redundancy.}
\label{fig:detailed_modules2}
\end{figure*}

\textit{2) Region Dynamic Convolution: }Guided by the partition matrix $\bm{M}$, this block adaptively generates distinct convolution kernels \cite{28} for each self-similar partition to achieve spatial adaptability. It is theoretically possible to use a simple multi-layer perceptron to directly map the cluster centroid to the full convolution weights. However, this naive approach entails an exorbitant number of learnable parameters, making the network exceedingly difficult to train. To significantly reduce the parameter burden while maintaining generative flexibility, we introduce a lightweight kernel generation scheme based on a low-rank matrix decomposition strategy.

Specifically, for the $i$-th cluster denoted by the spatial coordinate set $S_i$, its centroid vector $\bm{c}_i \in \mathbb{R}^{k^2 \times C_{in}}$ is computed by averaging the unfolded neighborhood features $\bm{p}_{x,y}$ of all pixels within $S_i$:
\begin{equation}\label{eq:centroid}
\bm{c}_i = \frac{1}{|S_i|} \sum_{(x,y) \in S_i} \bm{p}_{x,y}
\end{equation}

This centroid vector $\bm{c}_i$ is then flattened and fed into three parallel MLP branches. The top branch generates the bias vector $\bm{b}_i$. The middle and bottom branches generate two low-rank matrices, which are then multiplied to reconstruct the final convolution weights $\bm{W}_i$. The calculation process is formulated as:
\begin{subequations}\label{eq:adaptive_weights}
\begin{align}
\bm{A}_i &= \mathrm{MLP}_A(\bm{c}_i) \in \mathbb{R}^{(C_{in}\cdot k) \times rank} \\
\bm{B}_i &= \mathrm{MLP}_B(\bm{c}_i) \in \mathbb{R}^{rank \times (k\cdot C_{out})} \\
\bm{W}_i &= \bm{A}_i \bm{B}_i \in \mathbb{R}^{C_{in} \times k^2 \times C_{out}} \\
\bm{b}_i &= \mathrm{MLP}_{bias}(\bm{c}_i) \in \mathbb{R}^{C_{out}}
\end{align}
\end{subequations}

For each sub-band $\bm{P}_b \in \{\bm{P}_{LL}, \bm{P}_{LH}, \bm{P}_{HL}, \bm{P}_{HH}\}$, the partition-wise adaptive convolution is executed strictly within its corresponding cluster mask:
\begin{equation}\label{eq:pwac}
\bm{F}_{b}(x,y) = \bm{W}_{\bm{M}(x,y)} \otimes \bm{P}_{b}(x,y) + \bm{b}_{\bm{M}(x,y)}
\end{equation}

To ensure robust and deep spatial representation, the decoupled sub-bands are sequentially processed through a cascade of three identical SAC blocks. Following this progressive adaptive enhancement, the refined frequency representations must be projected back to the original spatial resolution. To accomplish this, the IDWT is applied. Functioning as the exact inverse mapping of Eq. \ref{eq:dwt_conv}, the IDWT implicitly combines the structured low-frequency representations with the adaptively amplified high-frequency details. Symmetrically, this is executed via fixed-weight transposed convolutions with a stride of 2 (denoted as $\uparrow_2$), guaranteeing mathematically precise and lossless spatial upsampling to produce the final enhanced feature map of the SAR module.

\subsection{CFA: Clustered Frequency Aggregation} \label{subsec:cfa}
To fuse information across frequencies while mitigating the heavy computational burden of traditional global attention, we propose the Clustered Frequency Aggregation (CFA) module. The CFA is composed of two phases: the Cluster-Routed Sparse Attention (CRSA) and the Gated Depthwise Forward Network (GDFN). Detailed architectures are shown in Fig. \ref{fig:detailed_modules2}.

\textit{1) Cluster-Routed Sparse Attention (CRSA): }Standard global self-attention incurs quadratic computational complexity and often integrates redundant interactions from semantically irrelevant regions. To overcome this limitation and improve computational efficiency, inspired by Roy et al. \cite{38}, the CRSA leverages the semantic clustering results from the SAR to execute a region-aware sparse attention mechanism. 

First, we construct a Frequency Attention Triplet comprising Frequency-Query, Spatial-Key, and Fusion-Value. The four frequency sub-bands derived from $\bm{P}_k$, denoted as $\bm{P}_b \in \{\bm{P}_{LL}, \bm{P}_{LH}, \bm{P}_{HL}, \bm{P}_{HH}\}$, act as the Frequency-Queries to capture specific frequency information. The low-frequency band $\bm{P}_{LL}$, which encapsulates the overall spatial appearance, serves as the Spatial-Key. Instead of directly using the raw LRMS image, the Fusion-Value is derived by fusing the current stage's intermediate multispectral features $\bm{M}_k$ with $\bm{P}_{LL}$. The generation of the triplet is formulated as:
\begin{subequations}\label{eq:triplet_gen}
\begin{align}
\bm{Q}_b &= \mathrm{MLP}(\mathrm{LN}(\bm{P}_b)) \\
\bm{K} &= \mathrm{MLP}(\mathrm{LN}(\bm{P}_{LL})) \\
\bm{V} &= \mathrm{MLP}(\mathrm{LN}(f_v(\bm{M}_k, \bm{P}_{LL})))
\end{align}
\end{subequations}
\noindent \lowercase{w}here $f_v$ denotes the convolutional fusion process, and $\mathrm{LN}$ represents Layer Normalization.

To significantly reduce redundancy and optimize computational cost, the attention score $A_b(i, j)$ for the query pixel at position $i$ and key pixel at position $j$ is computed via a bifurcated sparse logic routed by the cluster index $\bm{M}$ inherited from the SAR. For intra-cluster pixels, a strict pixel-to-pixel dot product is applied to retain dense, fine-grained correlations within the same semantic region. For inter-cluster pixels, the redundant pixel-wise calculations are discarded; instead, the relationship is modeled non-locally and sparsely using the cluster-mean key $\bar{\bm{K}}_c$. Specifically, $\bar{\bm{K}}_c$ is obtained by averaging the spatial-key embeddings of all pixels assigned to the $c$-th cluster:
\begin{equation}\label{eq:cluster_mean_key}
\bar{\bm{K}}_c = \frac{1}{|\Omega_c|} \sum_{j \in \Omega_c} \bm{K}_j, \quad \text{where} \quad \Omega_c = \{j \mid \bm{M}(j) = c\}
\end{equation}
where $|\Omega_c|$ denotes the number of pixels in this cluster. Thus, the bifurcated attention score $A_b(i, j)$ is formulated as:
\begin{equation}\label{eq:adaptive_attention}
A_b(i, j) = 
\begin{cases} 
\bm{Q}_{b,i} \bm{K}_j^T, & \text{if } \bm{M}(i) = \bm{M}(j) \\ 
\bm{Q}_{b,i} \bar{\bm{K}}_c^T, & \text{if } \bm{M}(i) \neq \bm{M}(j) 
\end{cases}
\end{equation}

By utilizing this sparse interaction pattern, the network focuses on effective features rather than global noise. The sparse attention map is then normalized via the Softmax function and aggregated with the Fusion-Value $\bm{V}$ to reconstruct the enhanced multi-frequency features $\bm{I}_b$:
\begin{equation}\label{eq:attention_agg}
\bm{I}_b = f_{out}\left( \mathrm{Softmax}\left(\frac{A_b}{\sqrt{d}}\right) \bm{V} \right)
\end{equation}
\noindent \lowercase{w}here $d$ is the scaling factor and $f_{out}$ denotes a MLP function. Subsequently, the IDWT is applied to the four reconstructed frequency features to obtain the spatial-domain representation $\bm{F}_{CRSA} = \mathrm{IDWT}(\bm{I}_{LL}, \bm{I}_{LH}, \bm{I}_{HL}, \bm{I}_{HH})$.

\textit{2) Gated Depthwise Forward Network (GDFN): }To further suppress redundant information passing through the sparse attention mechanism, $\bm{F}_{CRSA}$ is refined by the GDFN \cite{46}. This network implements a gating mechanism through parallel depthwise convolutions and the GELU activation:
\begin{subequations}\label{eq:gdfn}
\begin{align}
&\bm{X}_1 = \mathrm{GELU}\big(\phi_1(\mathrm{LN}(\bm{X}))\big) \odot \phi_2(\mathrm{LN}(\bm{X})) \label{eq:gdfn_a} \\
&\mathrm{GDFN}(\bm{X}) = \bm{X} + \mathrm{Proj}(\bm{X}_1) \label{eq:gdfn_b}
\end{align}
\end{subequations}
\noindent \lowercase{w}here $\phi_1$ and $\phi_2$ denote $3\times 3$ depthwise convolutions, $\mathrm{LN}$ represents Layer Normalization, $\odot$ denotes element-wise multiplication, and $\mathrm{Proj}$ is a $1\times 1$ linear projection.

To further refine the fused features and capture deeper representations, the output of the GDFN is subsequently fed into a cascaded CRSA block. However, unlike the initial CRSA which directly inherits the semantic cluster mask $\bm{M}$ from the SAR module, the intermediate low-frequency (LL) components at this cascaded stage have not undergone spatial clustering. To support the region-aware routing mechanism in this subsequent attention block, an independent Cluster Region Partition (CRP) operation is explicitly executed on these updated LL features.

\subsection{Network Framework and Loss Function}

To fully harness the multi-scale characteristics of remote sensing images, our overall network is formulated as a hierarchical framework with $N$ cascading scales. We first construct a wavelet pyramid by continuously applying the DWT operation to the feature representations. Let $\bm{P}_k$ denote the frequency features at the $k$-th scale; the decomposition from a larger scale to a smaller scale is formulated as:
\begin{equation}\label{eq:wavelet_pyramid}
\bm{P}_k = \mathrm{DWT}(\bm{P}_{k+1})
\end{equation}
\noindent \lowercase{w}here $k \in \{1, 2, \dots, N-1\}$. The spatial resolution progressively decreases as $k$ decreases.

The fusion paradigm operates progressively, starting from the lowest resolution scale and climbing up to the full scale. At each specific scale $k$, the network incorporates both the intermediate fused feature $\bm{M}_k$ and the panchromatic frequency features $\bm{P}_k$. The coupled operations of the aforementioned SAR and CFA process these inputs jointly. By denoting the synergistic function of these modules as $\mathcal{F}(\cdot)$, the progressive reconstruction step can be generalized as:
\begin{equation}\label{eq:progressive_fusion}
\bm{M}_{k+1} = \mathcal{F}(\bm{M}_k, \bm{P}_k)
\end{equation}

By recursively unrolling this iterative refinement across all $N$ scales, the deep representation at the final scale, $\bm{M}_N$, encapsulates rich spectral-spatial correspondences. A subsequent convolutional layer maps $\bm{M}_N$ to generate the definitive high-resolution multispectral image $\hat{\bm{I}}_{HRMS}$.

For the optimization of the entire RAF network, we employ a straightforward yet highly effective objective function. To strictly constrain the pixel-wise discrepancies and maintain stable convergence, the network is trained end-to-end using the standard $\ell_1$ loss function between the network's prediction $\hat{\bm{I}}_{HRMS}$ and the corresponding ground truth $\bm{I}_H$.

\begin{table}[!t]
\caption{Details of the two datasets (GaoFen2 and WorldView-3) used in this study}
\label{tab:tab1}
\centering
\fontsize{6pt}{8pt}\selectfont
\begin{adjustbox}{center, max width=\textwidth}
\begin{tabular}{>{\centering}cccccc}
\toprule
Datasets & Train & Reduced-resolution Test & Full-resolution Test \\
\midrule
GaoFen-2 & \makecell{Patches: 19809\\MS: $16 \times 16 \times 4$\\PAN: $64 \times 64 \times 1$} & \makecell{Patches: 20\\MS: $64 \times 64 \times 4$\\PAN: $256 \times 256 \times 1$} & \makecell{Patches: 20\\MS: $128 \times 128 \times 4$\\PAN: $512 \times 512 \times 1$} \\
\cmidrule(lr){1-4}
WorldView-3 & \makecell{Patches: 9714\\MS: $16 \times 16 \times 8$\\PAN: $64 \times 64 \times 1$} & \makecell{Patches: 20\\MS: $16 \times 16 \times 8$\\PAN: $64 \times 64 \times 1$} & \makecell{Patches: 20\\MS: $64 \times 64 \times 8$\\PAN: $256 \times 256 \times 1$} \\
\bottomrule
\end{tabular}
\end{adjustbox}
\label{tab:datasets}
\end{table}

\section{Experiments}
In this section, we summarize the datasets, implementation details, and experimental results. We conduct a series of comprehensive experiments to compare the fusion performance of our method against current state-of-the-art approaches. Furthermore, ablation studies are provided to validate the effectiveness of our proposed modules and architecture.

\subsection{Datasets and Benchmarks}
To evaluate the effectiveness of our network in pansharpening, we utilized multiple datasets acquired by sensors including WorldView-3 (WV3) and GaoFen-2 (GF2). Since ground-truth (GT) HRMS images are unavailable in real-world scenarios, we adopted Wald's protocol \cite{47} for simulated experiments. Each training dataset comprises pairs of PAN, LRMS, and GT images; detailed information is provided in Table~\ref{tab:tab1}. To benchmark the proposed RAF network, we selected several state-of-the-art pansharpening methods, including two traditional methods (GS \cite{48}, BDSD \cite{49}) and seven deep learning-based methods (PanNet \cite{50}, GPPNN \cite{51}, HyperTransformer \cite{45}, FAME \cite{52}, DCPNet \cite{53}, and FSGformer \cite{31}).

For reduced-resolution evaluations, we utilized common quantitative metrics, including Peak Signal-to-Noise Ratio (PSNR), Structural Similarity (SSIM), Spatial Correlation Coefficient (SCC), Spectral Angle Mapper (SAM), and the Erreur Relative Globale Adimensionnelle de Synthèse (ERGAS) \cite{54}. For full-resolution evaluations, we employed no-reference metrics including the Hybrid Quality with No Reference (HQNR), the spectral distortion index $D_\lambda$, and the spatial distortion index $D_s$ \cite{55}.

\begin{table*}[!t]
\renewcommand{\arraystretch}{1.3} 
\caption{Quantitative results of the fusion model on the WorldView-3 Reduced-resolution dataset. Bold shows the best result, marked with an underline shows for second-best results. Methods the number represents around the related papers published year.}
\label{tab:tab2} 
\centering 
\begin{tabular}{>{\centering}cccccc} 
\toprule
\textbf{Method} & \textbf{PSNR↑} & \textbf{SSIM↑} & \textbf{SCC↑} & \textbf{SAM↓} & \textbf{ERGAS↓} \\ 
\midrule
\textbf{GS}\cite{48}$_{2000}$             & 29.3142 ± 1.9269 & 0.8513 ± 0.0330 & 0.9238 ± 0.0343 & 6.1275 ± 1.7823 & 5.5237 ± 1.3956 \\
\textbf{BDSD}\cite{49}$_{2007}$           & 30.9460 ± 1.4069 & 0.8882 ± 0.0295 & 0.9352 ± 0.0276 & 5.4651 ± 1.6705 & 4.6536 ± 1.4292 \\
\textbf{PanNet}\cite{50}$_{2017}$         & 35.0838 ± 1.7267 & 0.9583 ± 0.0113 & 0.9721 ± 0.0257 & 3.7926 ± 0.7145 & 2.8359 ± 0.7002 \\
\textbf{GPPNN}\cite{51}$_{2021}$          & 34.3080 ± 1.5424 & 0.9534 ± 0.0116 & 0.9666 ± 0.0240 & 4.2531 ± 0.8029 & 3.0810 ± 0.7010 \\
\textbf{HyperTransformer}\cite{45}$_{2022}$& 36.1352 ± 1.6314 & 0.9649 ± 0.0092 & 0.9763 ± 0.0157 & 3.3702 ± 0.4964 & 2.4967 ± 0.5977 \\
\textbf{FAME}\cite{52}$_{2024}$           & 34.3259 ± 1.7426 & 0.9476 ± 0.0130 & 0.9633 ± 0.0366 & 4.2500 ± 0.7612 & 3.1115 ± 0.5954 \\
\textbf{DCPNet}\cite{53}$_{2024}$         & 36.9485 ± 1.7364 & 0.9679 ± 0.0087 & 0.9786 ± 0.0157 & 3.1042 ± 0.5343 & 2.2942 ± 0.4899 \\
\textbf{FSGformer}\cite{31}$_{2025}$      & \underline{37.9639 ± 1.7737} & \underline{0.9759 ± 0.0068} & \underline{0.9845 ± 0.0131} & \underline{2.7875 ± 0.5033} & \underline{2.0353 ± 0.4401} \\
\midrule
\textbf{RAFNet} (Ours)                    & \textbf{38.3770 ± 1.6536} & \textbf{0.9775 ± 0.0064} & \textbf{0.9857 ± 0.0126} & \textbf{2.5386 ± 0.5638} & \textbf{1.9687 ± 0.4368} \\
\bottomrule
\end{tabular}
\end{table*}

\begin{table*}[!t]
\renewcommand{\arraystretch}{1.3}
\caption{Quantitative results of the fusion model on the GaoFen-2 Reduced-Resolution dataset. Bold shows the best result, marked with an underline for second-best method. Methods the number represents around the related papers published year.}
\label{tab:tab3} 
\centering 
\begin{tabular}{>{\centering}cccccc} 
\toprule
\textbf{Method} & \textbf{PSNR↑} & \textbf{SSIM↑} & \textbf{SCC↑} & \textbf{SAM↓} & \textbf{ERGAS↓} \\ 
\midrule
\textbf{GS}\cite{48}$_{2000}$ & 31.5261 ± 1.7524 & 0.8838 ± 0.0327 & 0.9297 ± 0.0260 & 2.0877 ± 0.3819 & 2.4199 ± 0.4046 \\
\textbf{BDSD}\cite{49}$_{2007}$ & 34.9580 ± 2.0107 & 0.9029 ± 0.0299 & 0.9644 ± 0.0177 & 1.7246 ± 0.3118 & 1.6952 ± 0.3896 \\
\textbf{PanNet}\cite{50}$_{2017}$ & 38.8802 ± 1.8542 & 0.9578 ± 0.0101 & 0.9851 ± 0.0070 & 1.1240 ± 0.1977 & 1.7705 ± 0.2998 \\
\textbf{GPPNN}\cite{51}$_{2021}$ & 36.3356 ± 1.7038 & 0.9419 ± 0.0135 & 0.9733 ± 0.0116 & 1.4236 ± 0.2248 & 1.0825 ± 0.2028 \\
\textbf{HyperTransformer}\cite{45}$_{2022}$ & 41.7755 ± 1.6534 & 0.9761 ± 0.0052 & 0.9923 ± 0.0036 & 0.8168 ± 0.1462 & 0.7632 ± 0.1261 \\
\textbf{FAME}\cite{52}$_{2024}$ & 40.2324 ± 1.5014 & 0.9709 ± 0.0053 & 0.9900 ± 0.0048 & 0.8751 ± 0.1548 & 0.8947 ± 0.1468 \\
\textbf{DCPNet}\cite{53}$_{2024}$ & 42.7252 ± 1.4202 & 0.9813 ± 0.0035 & 0.9940 ± 0.0029 & 0.7709 ± 0.0987 & 0.6861 ± 0.0844 \\
\textbf{FSGformer}\cite{31}$_{2025}$ & \underline{45.0504 ± 1.4875} & \underline{0.9873 ± 0.0027} & \underline{0.9962 ± 0.0019} & \underline{0.5968 ± 0.1008} & \underline{0.5201 ± 0.0816} \\
\midrule 
\textbf{RAFNet} (Ours) & \textbf{45.1372 ± 1.4356} & \textbf{0.9892 ± 0.0026} & \textbf{0.9968 ± 0.0018} & \textbf{0.5759 ± 0.0936} & \textbf{0.5001 ± 0.0796} \\
\bottomrule
\end{tabular}
\end{table*}

\subsection{Implementation Details}
The proposed RAFNet was implemented in PyTorch and trained on an NVIDIA RTX 5090 GPU. We utilized the Adam optimizer with an initial learning rate of $9 \times 10^{-4}$, halved every 90 epochs, for a total of 500 epochs (batch size 32). In the CRSA module, the number of attention heads was set to 8. The semantic cluster number $K$ was configured to 16 during training and scaled to 64 during testing to accommodate the larger spatial dimensions and diverse frequency distributions of full-scene images.

\subsection{Reduced-Resolution Assessment}
\textit{1) Quantitative Analysis: }Tables \ref{tab:tab2} and \ref{tab:tab3} report the quantitative results obtained by different methods on the WorldView-3 and GaoFen-2 reduced-resolution datasets, respectively. Deep learning-based methods consistently outperform traditional approaches (GS and BDSD), underscoring the powerful representation capabilities of neural networks. Notably, the proposed RAFNet achieves state-of-the-art performance on both datasets, significantly outperforming the second-best method (FSGformer) across all metrics. These consistent improvements can be attributed to the synergistic integration of the SAR and CFA modules. The SAR module ensures precise directional frequency decoupling and regional feature enhancement, while the CFA module effectively captures long-range dependencies, jointly ensuring superior spatial fidelity and spectral consistency.

\begin{figure*}[!t]
  \centering  
  \hypertarget{fig:fig7}{}
  \includegraphics[width=0.9\textwidth]{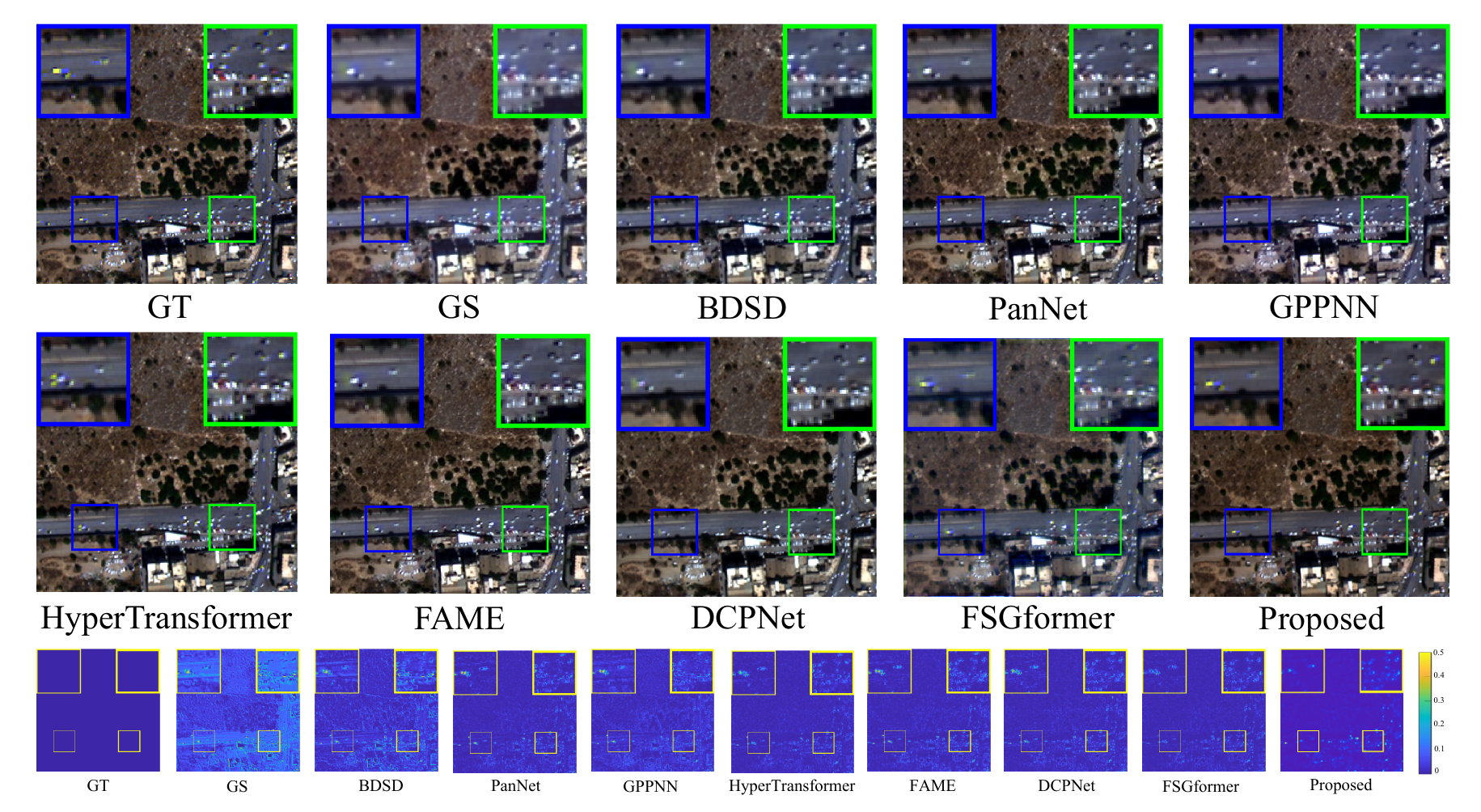}
  \caption{The visual results (Top) and residuals (Bottom) of all compared approaches on the WorldView-3 reduced-resolution dataset.}
  \label{fig:fig7}
\end{figure*}

\begin{figure*}[!t]
  \centering  
  \hypertarget{fig:fig8}{}
  \includegraphics[width=0.9\textwidth]{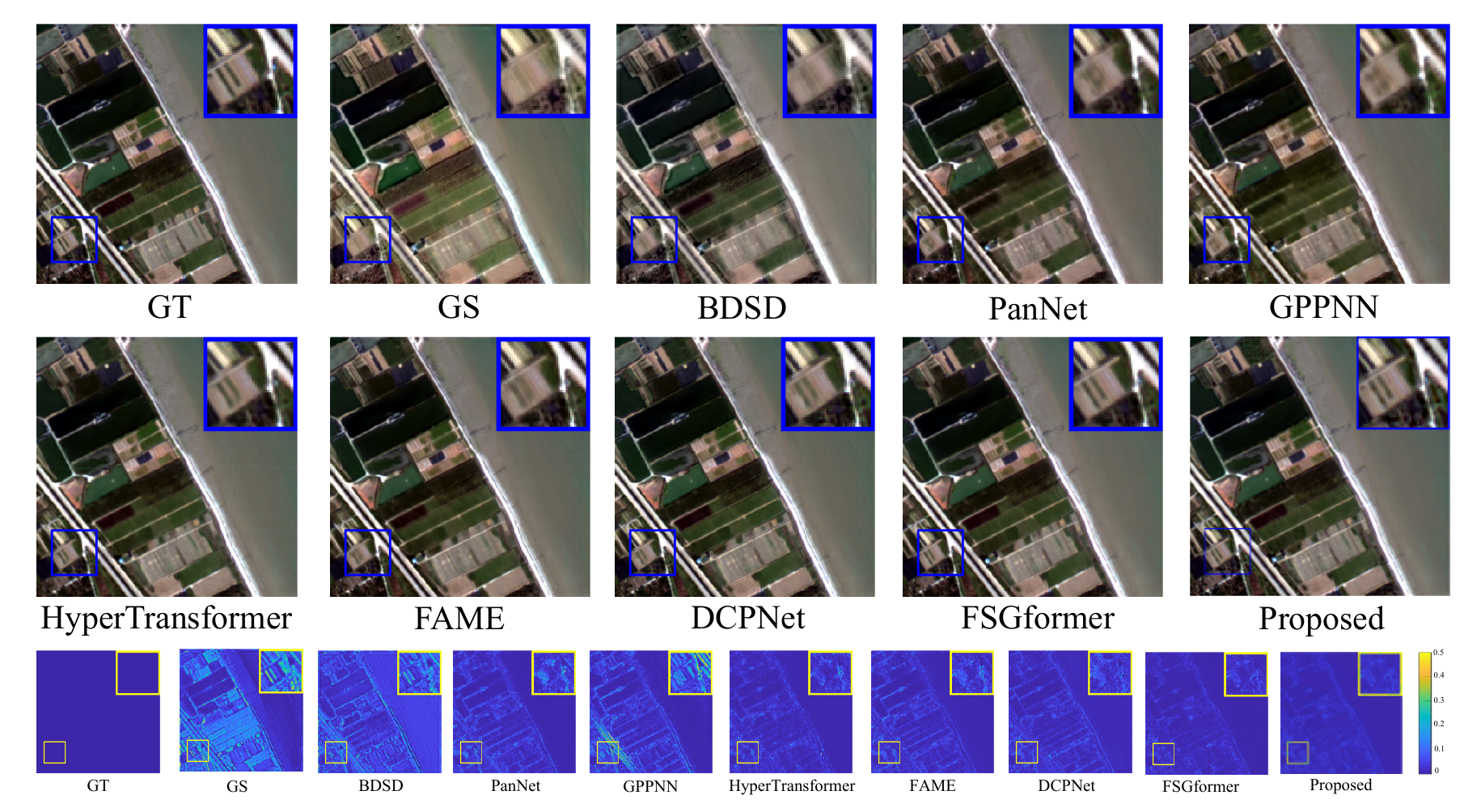}
  \caption{The visual results (Top) and residuals (Bottom) of all compared approaches on the Gaofen-2 reduced-resolution dataset.}
  \label{fig:fig8}
\end{figure*}

\textit{2) Qualitative Analysis: }Figures \ref{fig:fig7} and \ref{fig:fig8} present visual comparisons and residual maps on the reduced-resolution datasets. As observed in the magnified regions, traditional methods and several earlier deep learning models exhibit blurred structures and edge artifacts. In contrast, RAFNet produces sharper contours and finer spatial details that are most consistent with the ground-truth image. The corresponding residual maps further support this observation, where the residuals of RAFNet are noticeably closer to blue (representing zero error) across the entire scene, indicating higher reconstruction accuracy.

\subsection{Full-Resolution Assessment}
\textit{1) Quantitative Analysis: }To further verify practical performance, we evaluate all methods on the full-resolution datasets without reference images (Table \ref{tab:tab4}). RAFNet achieves the best performance across all three no-reference metrics (HQNR, $D_\lambda$, and $D_s$) for both sensors. The lowest distortion indices ($D_\lambda$ and $D_s$) indicate that our method effectively preserves the spectral information of the LRMS image while successfully injecting high-resolution spatial details from the PAN image. The highest HQNR values further demonstrate that RAFNet maintains excellent generalization ability and superior overall quality in real-world scenarios where ground-truth references are unavailable.

\begin{table}[!t]
\caption{Quantitative results of the fusion model on WorldView-3 and GaoFen-2 datasets. The best results are highlighted in bold, and the underlined ones are the second-best methods. The numbers around the methods represent the publication year of the relevant paper.}
\label{tab:tab4}
\centering
\fontsize{7.5pt}{10pt}\selectfont
\setlength{\tabcolsep}{1.8pt} 
\begin{tabular}{>{\centering}c ccc ccc}
\toprule
\multirow{2}{*}{\textbf{Method}} & \multicolumn{3}{c}{\textbf{WorldView-3}} & \multicolumn{3}{c}{\textbf{GaoFen-2}} \\
\cmidrule(lr){2-4} \cmidrule(lr){5-7}
& \textbf{HQNR↑} & \textbf{$D_\lambda$↓} & \textbf{$D_s$↓} & \textbf{HQNR↑} & \textbf{$D_\lambda$↓} & \textbf{$D_s$↓} \\
\midrule
\textbf{GS}\cite{48}$_{2000}$ & 0.8492 & 0.0929 & 0.0646 & 0.6216 & 0.2176 & 0.2077 \\
\textbf{BDSD}\cite{49}$_{2007}$ & 0.8698 & 0.0625 & 0.0730 & 0.7812 & 0.0759 & 0.1548 \\
\textbf{PanNet}\cite{50}$_{2017}$ & 0.8760 & 0.0586 & 0.0703 & 0.8962 & 0.0298 & 0.0763 \\
\textbf{GPPNN}\cite{51}$_{2021}$ & 0.8507 & 0.0750 & 0.0814 & 0.8553 & 0.0536 & 0.0963 \\
\textbf{HyperTransformer}\cite{45}$_{2022}$ & 0.9463 & 0.0217 & 0.0328 & 0.9269 & 0.0292 & 0.0451 \\
\textbf{FAME}\cite{52}$_{2024}$ & 0.9158 & 0.0312 & 0.0549 & 0.9299 & 0.0235 & 0.0486 \\
\textbf{DCPNet}\cite{53}$_{2024}$ & 0.9196 & 0.0412 & 0.0410 & 0.9165 & 0.0258 & 0.0592 \\
\textbf{FSGformer}\cite{31}$_{2025}$ & \underline{0.9598} & \underline{0.0165} & \underline{0.0241} & \underline{0.9580} & \underline{0.0186} & \underline{0.0238} \\
\midrule
\textbf{RAFNet} (Ours) & \textbf{0.9610} & \textbf{0.0142} & \textbf{0.0213} & \textbf{0.9615} & \textbf{0.0171} & \textbf{0.0223} \\
\bottomrule
\end{tabular}
\end{table}

\begin{figure*}[!t]
  \centering  
  \hypertarget{fig:fig9}{}
  \includegraphics[width=1\textwidth]{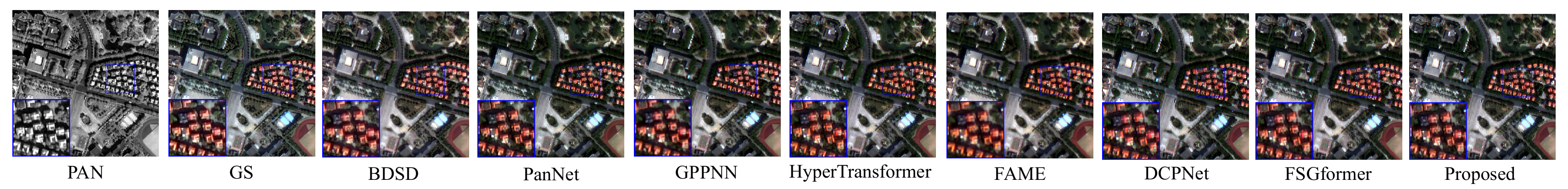}
  \caption{The visual results of all compared approaches on the GaoFen-2 full-resolution dataset.}
  \label{fig:fig9}
\end{figure*}

\textit{2) Qualitative Analysis: }
Fig. \ref{fig:fig9} presents the qualitative results on the GaoFen-2 full-resolution dataset. For a more comprehensive comparison, the corresponding PAN image is also provided, and several representative regions are enlarged for detailed inspection. As observed in the magnified areas, the proposed method produces more natural visual appearances with clearer structural details and better spectral consistency compared with other methods. This indicates that the proposed framework effectively captures and integrates spatial and spectral information to generate high-quality fusion results.

\subsection{Computational Complexity Analysis}
To evaluate computational efficiency, we compare the total learnable parameters (Params) and floating-point operations (FLOPs) using a simulated $256 \times 256$ PAN image and a $64 \times 64 \times C$ LRMS image. Furthermore, to explicitly highlight the efficiency of our proposed attention mechanism, we separately report the parameters specifically allocated to the Transformer or attention modules (Trans. Params) for the relevant network architectures. 

As detailed in Table \ref{tab:complexity}, early CNNs are extremely lightweight but yield suboptimal fusion performance due to their limited receptive fields. Standard Transformers achieve high spatial and spectral fidelity but suffer from significant computational burdens and parameter inflation. This inflation is primarily driven by the massive linear projection matrices required for the quadratic $\mathcal{O}(N^2)$ global self-attention mechanism and their associated heavy feed-forward networks. For instance, the attention-related modules in FSGformer alone consume the vast majority of its overall parameter budget.

In contrast, RAFNet achieves an optimal trade-off between performance and efficiency. By utilizing the CRSA module, dense dot-product computations are restricted to localized semantic clusters $\mathcal{O}(\frac{N^2}{K})$, while inter-cluster dependencies are efficiently modeled via lightweight cluster-mean keys $\mathcal{O}(N \times K)$. This region-aware routing strategy circumvents the need for massive, redundant projection weights typical in standard global attention. Consequently, the attention component of RAFNet operates with a substantially smaller parameter footprint compared to other Transformer-based counterparts. Overall, RAFNet effectively curtails both FLOPs and parameter redundancy while maintaining superior representation capability, establishing a highly efficient fusion paradigm.

\begin{table}[!t]
\caption{Comparison of computational complexity among different deep learning-based methods, including the parameter count specifically allocated to attention/Transformer modules. Evaluated on a $256 \times 256$ PAN image.}
\label{tab:complexity}
\centering
\fontsize{8pt}{10pt}\selectfont
\setlength{\tabcolsep}{3pt} 
\begin{adjustbox}{max width=\columnwidth} 
\begin{tabular}{>{\centering\arraybackslash}m{3.2cm} ccc} 
\toprule
\textbf{Method} & \makecell{\textbf{Total Params} \\ \textbf{(M)↓}} & \makecell{\textbf{Trans. Params} \\ \textbf{(M)↓}} & \textbf{FLOPs (G)↓} \\
\midrule
\textbf{PanNet}\cite{50} & 0.07 & - & 0.12 \\
\textbf{GPPNN}\cite{51} & 0.15 & - & 0.35 \\
\textbf{HyperTransformer}~\cite{45} & 5.45 & 4.21 & 3.52 \\ 
\textbf{FAME}\cite{52} & 2.85 & - & 1.85 \\
\textbf{DCPNet}\cite{53} & 3.10 & 2.95 & 4.35 \\
\textbf{FSGformer}\cite{31} & 9.80 & 8.62 & 2.18 \\
\midrule
\textbf{RAFNet (Ours)} & \textbf{3.09} & \textbf{2.92} & \textbf{1.38} \\
\bottomrule
\end{tabular}
\end{adjustbox}
\end{table}

\subsection{Ablation Study}
To validate the proposed components, we conducted comprehensive ablation studies on the WorldView-3 dataset, with quantitative results summarized in Table \ref{tab:ablation_merged}.

\subsubsection{Attention Mechanism Analysis}
We evaluate the superiority of our CRSA by replacing it with three alternative masking strategies while keeping the rest of the network architecture unchanged. Specifically, the first strategy is Local Block Attention, which restricts query-key interactions to non-overlapping $4 \times 4$ local spatial windows. The second strategy is Banded Attention, which constrains attention to a local diagonal band (bandwidth = 8) within the flattened sequence. The third strategy is Global Cross Attention, which computes sparse global attention utilizing a cross-shaped mask (cross-width = 32) to capture long-range interactions along the horizontal and vertical axes.

As shown in Table \ref{tab:ablation_merged}, Local and Banded attentions yield suboptimal performance by rigidly severing long-range dependencies. Conversely, while Global Cross expands the receptive field, it indiscriminately incorporates semantically irrelevant pixels, leading to noise aggregation and structural blurring. In contrast, the proposed CRSA dynamically routes interactions via K-means clusters. This strategy effectively filters out redundant noise while capturing highly relevant non-local dependencies, achieving the highest performance across all metrics.

\subsubsection{Network Module Analysis}
To verify the complementary nature of the SAR and CFA modules, we evaluated two constrained configurations: Only-Frequency (removing SAR) and Only-Spatial (removing CFA). Removing the SAR module results in a noticeable performance drop, highlighting that purely frequency-domain attention struggles to capture fine-grained, localized spatial variations without adaptive convolutions. Conversely, removing the CFA module restricts the network to localized operations, severely limiting its ability to model global contextual information and cross-frequency correlations. The superior performance of the full RAFNet demonstrates that the spatial adaptability of the SAR and the non-local frequency routing of the CFA are mutually indispensable for high-fidelity pansharpening.

\begin{table}[!t]
\caption{Quantitative results of ablation studies on the WorldView-3 dataset. The best results are highlighted in bold.}
\centering
\fontsize{8pt}{10pt}\selectfont
\setlength{\tabcolsep}{4pt} 
\begin{adjustbox}{center, max width=0.95\textwidth}
\begin{tabular}{>{\centering\arraybackslash}l ccccc}
\toprule
\textbf{Configuration} & \textbf{PSNR↑} & \textbf{SSIM↑} & \textbf{SCC↑} & \textbf{SAM↓} & \textbf{ERGAS↓} \\
\midrule
\multicolumn{6}{c}{\textit{Attention Mechanism Variants}} \\
\midrule
Local Block Attention  & 38.0666 & 0.9715 & 0.9785 & 2.8941 & 2.2014 \\
Banded Attention       & 38.1720 & 0.9732 & 0.9810 & 2.8023 & 2.1252 \\
Global Cross Attention & 38.1992 & 0.9745 & 0.9821 & 2.7658 & 2.0854 \\
\textbf{CRSA (Ours)}   & \textbf{38.3770} & \textbf{0.9775} & \textbf{0.9857} & \textbf{2.5386} & \textbf{1.9687} \\
\midrule
\multicolumn{6}{c}{\textit{Network Module Configurations}} \\
\midrule
Only-Spatial (w/o CFA)  & 38.2377 & 0.9748 & 0.9825 & 2.7139 & 2.0578 \\
Only-Frequency (w/o SAR)& 38.2824 & 0.9758 & 0.9835 & 2.6512 & 2.0215 \\
\textbf{RAFNet (Full)}  & \textbf{38.3770} & \textbf{0.9775} & \textbf{0.9857} & \textbf{2.5386} & \textbf{1.9687} \\
\bottomrule
\end{tabular}
\end{adjustbox}
\label{tab:ablation_merged}
\end{table}

\begin{figure*}[!t]
  \centering
  \includegraphics[width=0.9\textwidth]{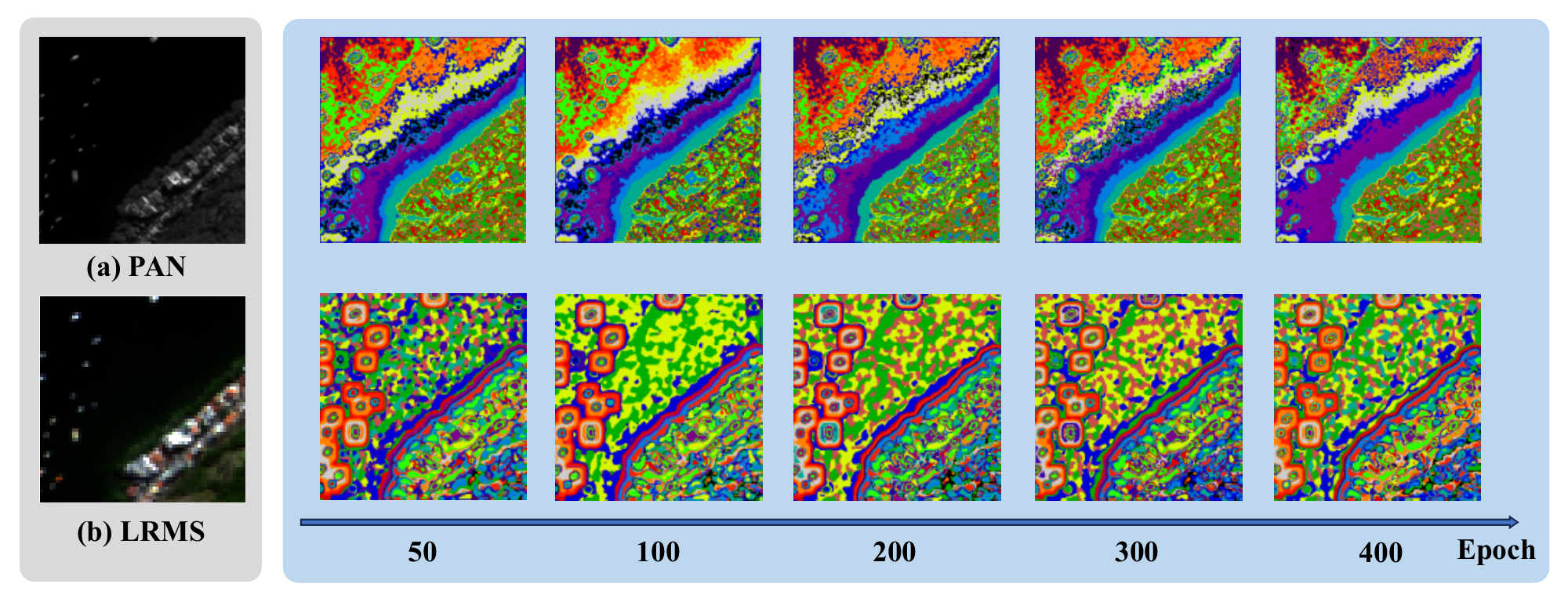}
  \caption{Visual representations of cluster index matrices in the model at different training epochs. (a) and (b) show the visual appearance of the raw PAN and LRMS input images. The right section displays the clustering results on these raw inputs without being transformed by convolution layers, where the upper row corresponds to the PAN image and the lower row corresponds to the LRMS image. The color indicates the cluster to which the pixel belongs.}
  \label{fig:ablation_visual}
\end{figure*}

\begin{figure*}[!t]
  \centering
  \includegraphics[width=0.75\textwidth]{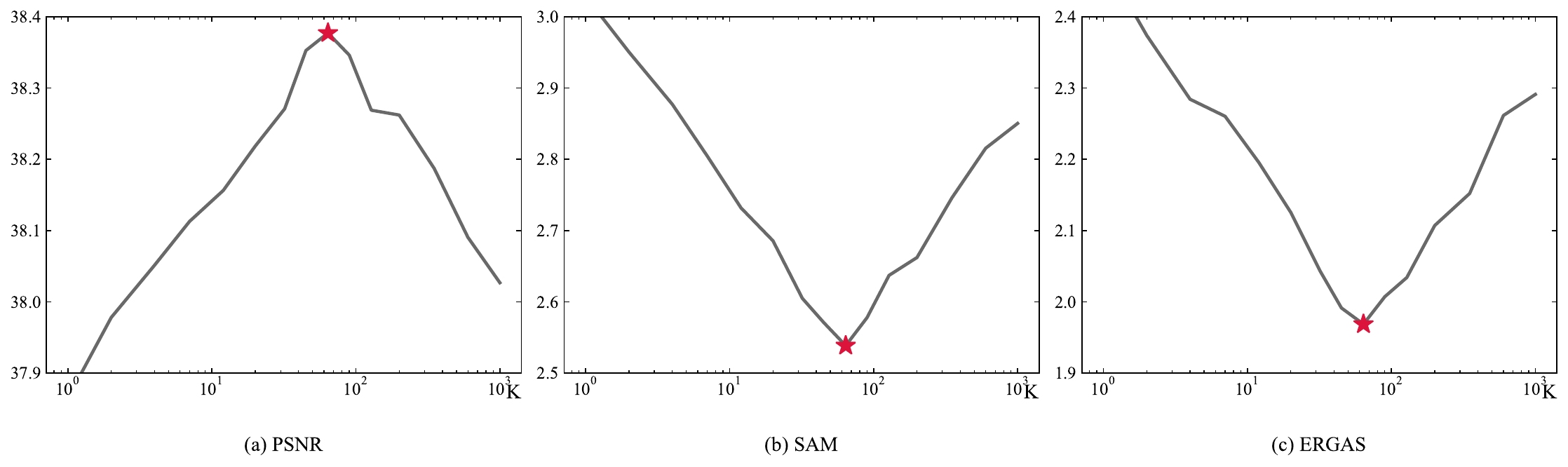}
  \caption{Variations of PSNR, SAM and ERGAS on the WorldView-3 reduced-resolution dataset with changing cluster number $K$. The optimal metrics are obtained around $K = 64$.}
  \label{fig:ablation_k}
\end{figure*}

\subsubsection{Analysis of Semantic Clustering}
To comprehensively evaluate the clustering mechanism within the SAR module, we analyze both the visual morphology of the partitions (Fig. \ref{fig:ablation_visual}) and the sensitivity of the cluster count $K$ (Fig. \ref{fig:ablation_k}). Visually, as training converges, the SAR module progressively transitions from grouping low-level color features to jointly modeling non-local semantic similarities. Homogeneous regions aggregate into continuous clusters, while high-frequency details are isolated into distinct subsets, ensuring that subsequent dynamic convolutions and sparse attention operate on precise semantic regions.

Quantitatively, the performance metrics in Fig. \ref{fig:ablation_k} demonstrate that appropriate clustering is crucial for non-local information transmission. A small $K$ approximates global filtering by improperly grouping dissimilar pixels, whereas an overly large $K$ generates excessively sparse clusters that impede robust feature aggregation. Notably, the optimal $K$ during inference (64) is larger than the training $K$ (16), which naturally compensates for the larger spatial extent and more diverse frequency distributions inherent in full-scene test images.

\bibliographystyle{IEEEtran}
\bibliography{IEEEabrv,references}

\end{document}